\documentclass[letterpaper, 10 pt, conference]{ieeeconf}  

\IEEEoverridecommandlockouts                              

\usepackage{ntheorem}
\usepackage{booktabs} 
\usepackage{adjustbox} 
\usepackage{cite}
\usepackage{amsmath,amssymb,amsfonts}
\usepackage{algorithmic}
\usepackage{graphicx}
\usepackage{textcomp}
\usepackage{xcolor}
\usepackage{hyperref}
\newtheorem{theorem}{Theorem} 
\newcommand{\norm}[2]{\left \lVert #1 \right \rVert_{#2}}

\title{\LARGE \bf
A Mixed-Integer Conic Program for the Multi-Agent Moving-Target Traveling Salesman Problem
}

\author{Allen George Philip$^{1}$, Zhongqiang Ren$^{2}$, Sivakumar Rathinam$^{3}$ and Howie Choset$^{2}$
    \thanks{$^{1}$ Mechanical Engineering, Texas A\&M University,
		College Station, TX 77843-3123.
		Email: {\tt y262u297@tamu.edu}}%
    \thanks{$^{2}$Robotics Institute, Carnegie Mellon University, 5000 Forbes Ave., Pittsburgh, PA 15213, USA.}
    \thanks{$^{3}$Mechanical Engineering, and Computer Science and Engineering, Texas A\&M University,
		College Station, TX 77843-3123.
}
}

\begin{document}

\maketitle
\thispagestyle{empty}
\pagestyle{empty}

\begin{abstract}
The Moving-Target Traveling Salesman Problem (MT-TSP) seeks a shortest path for an agent that starts at a stationary depot, visits a set of moving targets exactly once, each within one of their respective time windows, and returns to the depot. In this paper, we introduce a new Mixed-Integer Conic Program (MICP) formulation for the Multi-Agent Moving-Target Traveling Salesman Problem (MA-MT-TSP), a generalization of the MT-TSP involving multiple agents. Our approach begins by restating the current state-of-the-art MICP formulation for MA-MT-TSP as a Nonconvex Mixed-Integer Nonlinear Program (MINLP), followed by a novel reformulation into a new MICP. We present computational results demonstrating that our formulation outperforms the state-of-the-art, achieving up to two orders of magnitude reduction in runtime, and over 90\% improvement in optimality gap.

\end{abstract}

\vspace{2mm}
\section{Introduction}
Given a set of fixed target locations (targets) and the traversal costs between all target pairs, the Traveling Salesman Problem (TSP) aims to find a tour of minimum cost for an agent, that visits all the targets exactly once. The TSP is a classical problem in combinatorial optimization, with several applications including unmanned vehicle planning \cite{oberlin2010today, liu2018efficient, ryan1998reactive, yu2002implementation}, transportation and delivery \cite{ham2018integrated}, monitoring and surveillance \cite{venkatachalam2018two, saleh2004design}, disaster management \cite{cheikhrouhou2020cloud}, precision agriculture \cite{conesa2016mix}, and search and rescue \cite{zhao2015heuristic, brumitt1996dynamic}. The Moving-Target Traveling Salesman Problem (MT-TSP) is a generalization of the TSP where the targets traverse some predefined paths. The targets may also have associated time windows only during which an agent can visit them. Different variants of the MT-TSP were introduced in the literature, motivated by practical applications such as defending an area from oncoming hostile rockets or Unmanned Aerial Vehicles \cite{helvig2003, smith2021assessment, stieber2022}, monitoring and surveillance \cite{deMoraes2019, wang2023moving, marlow2007travelling, maskooki2023bi}, resupply missions with moving targets \cite{helvig2003}, dynamic target tracking \cite{englot2013efficient}, and industrial robot planning \cite{chalasani1999approximating}.

\vspace{1mm}
It is generally assumed that the maximum speed of the agent is greater than the speed of the fastest target \cite{helvig2003}. When the speeds of all the targets reduces to 0, the MT-TSP simplifies to the TSP. Hence, MT-TSP is NP-Hard. In the literature, we find several heuristic based approaches for the MT-TSP \cite{deMoraes2019, wang2023moving, marlow2007travelling, englot2013efficient, bourjolly2006orbit, choubey2013, groba2015solving, jiang2005tracking, ucar2019meta} that finds feasible solutions, but gives no information on how far they are from the optimum. The literature also presents exact and approximation algorithms for some MT-TSP variants. However, they only apply to very restricted cases where the targets are assumed to move in the same direction with the same speed \cite{chalasani1999approximating, hammar1999}, move along the same line \cite{helvig2003, hassoun2020}, or move along lines passing through the depot \cite{helvig2003}. In a recent paper \cite{philip2023c}, we presented an algorithm capable of handling targets moving along arbitrary curves, offering tight lower bounds for the MT-TSP. Also, in \cite{philip2024mixed}, we introduced a new mixed-integer conic program (MICP) that finds the optimum for the MT-TSP, for the case where each target moves along its own line, and have one associated time window. Finally, a complete algorithm for MT-TSP with stationary obstacles was also introduced recently in \cite{bhat2024complete}.

\begin{figure}
    \centering
    \includegraphics[width=\linewidth]{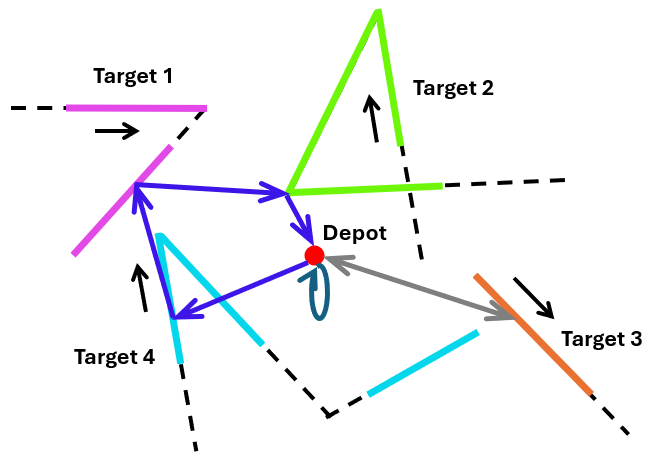}
    \caption{A feasible solution for an example instance of the MA-MT-TSP. The solid, colored portions of the target trajectories correspond to their time windows. The agents begin and end their tour at the depot. Note how one of the three agents is not assigned any targets in this solution, and simply waits at the depot.}
    \label{fig:mpmttsp}
    \vspace{-4mm}
\end{figure}

\vspace{1mm}
The objective of this paper is to find exact solutions for the Multi-Agent Moving-Target Traveling Salesman Problem (MA-MT-TSP), a generalization of the MT-TSP with multiple agents. Currently, the only exact solver for this problem is the MICP introduced in \cite{stieber2022}. Hence, we take this as our baseline. Here, each target is assumed to move along its own line, and have one associated time window. However, in this paper, we consider a less restricted case where each target traverse a path made of piecewise-linear segments, and have more than one associated time window (Fig.~\ref{fig:mpmttsp}). Therefore, we first extend the baseline MICP to accommodate these changes. Then, we pose this extended formulation as a nonconvex mixed-integer nonlinear program (nonconvex MINLP), which we then finally reformulate as a new MICP, similar to the works in \cite{philip2024mixed}, and \cite{marcucci2024shortest}.



\vspace{1mm}
We prove that our approach finds the optimum for the MA-MT-TSP, and present computational results to corroborate the performance of our formulation. We observe that our MICP formulation significantly outperforms the baseline, and scales much better with increasing number of targets and agents, and larger time window durations, achieving up to a two-order-of-magnitude reduction in runtime, and up to over 90\% tighter optimality gap. 

\section{Problem Definition}
All the targets and agents move on a 2D Euclidean plane ($\mathbb{R}^2$). All the agents share the same maximum speed $v_{max}$, and the same depot $s$. Without loss of generality, we define a copy of the depot, denoted by $s'$, and require the agents to return to $s'$ after completing their respective tours. Let $V_{tar} := \{1,2,\cdots, n\}$ denote the set of $n$ moving targets, and $V_{agt}:=\{1,2,\cdots,m\}$ denote the set of $m$ agents. For a target $u \in V_{tar}$, the portion of $u$'s trajectory within any of its time windows consists of linear trajectory segments. The set of all such segments\footnote{We represent a segment using a unique label $i \in \mathbb{Z}^+$.} across all of $u$'s time windows is denoted by $C_u$. The set $V_{seg} := \bigcup_{u\in V_{tar}}C_u$ consists of all the linear trajectory segments corresponding to all targets. For a segment $i \in V_{seg}$, $\tau_i \in V_{tar}$ denotes its associated target. Also, $\underline{t}_i$ and $\overline{t}_i$ represent the times, and $\underline{p}_i$ and $\overline{p}_i$ represent the positions of the segment's start and end points. Finally, $v_i$ denotes the velocity of $\tau_i$ along the segment. Let $V := V_{seg} \cup \{s,s'\}$. For $s$ and $s'$, we define velocities $v_{s}$, $v_{s'}$, and set them to $(0,0)$ since they are stationary. Also, we define time windows $[\underline{t}_s, \overline{t}_s]$ and $[\underline{t}_{s'}, \overline{t}_{s'}]$ for $s$ and $s'$ respectively. We set $\underline{t}_s = \overline{t}_s = 0$ since the agents depart from $s$ at time $0$. We also set $\underline{t}_{s'} = 0$ and $\overline{t}_{s'} = T$ where $T$ is the time horizon over which the trajectories of the targets are defined. For some $i \in \{s,s'\}$, $\underline{p}_i$ and $\overline{p}_i$ simply refers to the fixed depot location. An agent visits a moving target (say $u$) if the position of the agent coincides with the position of $u$ at some time $t$ within time interval $[\underline{t}_i, \overline{t}_i]$ corresponding to some segment $i \in C_u$. The objective of the MA-MT-TSP is to find tours for the agents such that each target is visited once by one of the agents, and the sum of the distances traversed by the agents is minimized.

\section{Baseline MICP for MA-MT-TSP}
This section presents the current state-of-the-art MICP for the MA-MT-TSP introduced in \cite{stieber2022}. We extend this formulation to accommodate piecewise-linear trajectories, and more than one time window for each target. Before proceeding further, we define a directed graph $(V,E)$, with all the nodes in $C_u$ for each $u\in V_{tar}$ forming a cluster. The edges in $E$ are added as follows: From $s$ to all nodes in $V_{seg}$, from each node belonging to a cluster to all the other nodes belonging to different clusters, and finally, from all the nodes in $V_{seg}$, to $s'$. For any node $i \in V$, $E_{i}^{in}$ and $E_{i}^{out}$ denotes the set of all edges entering and exiting $i$.

\vspace{1mm}
We now define the decision variables for the MICP. For each node $i \in V_{seg}$, the variable $t_{i,k} \in \mathbb{R}_{\geq 0}$ denotes the time at which an agent $k \in V_{agt}$ visits target $\tau_i$ at node $i$. The position of $\tau_i$ at time $t_{i,k}$ is denoted by $p_{i,k} \in \mathbb{R}^2$. Variables $t_{i,k}$ are also defined for $i \in \{s,s'\}$, denoting the time at which agent $k$ departs or arrives at the depot. Here, $p_{i,k}$ denotes the fixed depot position. For each edge $e=(i,j)\in E$, the binary variable $y_{e,k} \in \{0,1\}$ denotes the choice of whether agent $k$ traverses the edge. The auxiliary variable $l_{e,k} \in \mathbb{R}_{\geq 0}$ denotes the cost incurred by agent $k$ if it were to traverse the edge, which can be modeled by $y_{e,k}\norm{p_{j,k}-p_{i,k}}{2}$. To define each $l_{e,k}$ variable, the MICP relies on two other auxiliary variables $l_{e,k}^{xy} \in \mathbb{R}^2$, and $\overline{l}_{e,k} \in \mathbb{R}_{\geq 0}$. The variables $l_{e,k}^{xy}$ represent $p_{j,k}-p_{i,k}$, and variables $\overline{l}_{e,k}$ relate to $l_{e,k}$ through a set of big-$M$ constraints in the MICP. The big-$M$ here is the parameter $R$, which denotes the diagonal length of the square area that contains the depot and all the target trajectories. We now present the baseline formulation below.

\begin{align}
    \label{eq:SOCPobj}
    &\min \;\; \sum_{k\in V_{agt}}\sum_{e\in E} l_{e,k} \\
    \intertext{subject to constraints}
    \label{eq:depotFlowOut}
    &\sum_{e \in E^{out}_s}y_{e,k} \leq 1, \;\; \forall \; k \in V_{agt}, \\
    \label{eq:depotFlowIn}
    &\sum_{e \in E^{in}_{s'}}y_{e,k} \leq 1, \;\; \forall \; k \in V_{agt}, \\
    \label{eq:targetFlowIn}
    &\sum_{k\in V_{agt}}\sum_{i \in C_u}\sum_{e \in E^{in}_i}y_{e,k} = 1, \;\; \forall \; u \in V_{tar}, \\
    \label{eq:flowConservation}
    &\sum_{e \in E^{in}_i}y_{e,k} = \sum_{e \in E^{out}_i}y_{e,k}, \;\; \forall \; i \in V_{seg}, \; k \in V_{agt}, \\
    \label{eq:nodeTimeWin}
    & \underline{t}_i \leq t_{i,k} \leq \overline{t}_i, \;\; \forall \; i \in V, \; k \in V_{agt}, \\
    \label{eq:speedSOCPBigM}
    \begin{split}
        &l_{e,k} \leq v_{max}(t_{j,k}-t_{i,k}+T(1-y_{e,k})), \\
        &\forall \; e=(i,j) \in E, \; k \in V_{agt}, \\
    \end{split} \\
    \label{eq:defSOCPlxy}
    \begin{split}
        &l_{e,k}^{xy}=((\underline{p}_{j}+t_{j,k}v_{j}-\underline{t}_{j}v_{j}) -(\underline{p}_{i}+t_{i,k}v_{i}-\underline{t}_{i}v_{i})),\\
        &\forall \; e=(i,j) \in E, \; k \in V_{agt},
    \end{split} \\
    \label{eq:coneSOCPBigM}
    &\overline{l}_{e,k} = l_{e,k} + R(1-y_{e,k}), \; \forall \; e=(i,j) \in E, \; k \in V_{agt}, \\
    \label{eq:normSquaredSOCP}
    &\norm{l_{e,k}^{xy}}{2}^2 \leq \overline{l}_{e,k}^2, \;\; \forall \; e=(i,j) \in E, \; k \; \in V_{agt}.
\end{align}

\vspace{1mm}
The objective \eqref{eq:SOCPobj} is to minimize the sum of tour lengths of all the agents. The condition that an agent can depart from and arrive at the depot at most once is described by \eqref{eq:depotFlowOut} and \eqref{eq:depotFlowIn} respectively. The constraints \eqref{eq:targetFlowIn} ensure that each target is visited exactly once by one of the agents, at one of the target's segments. The flow conservation constraints for all the segment nodes, for all agents, are described by \eqref{eq:flowConservation}. Constraints \eqref{eq:depotFlowOut} to \eqref{eq:flowConservation} are flow constraints that ensure that each agent can have at most one tour, which starts and ends at the depot, and that each target must be visited once by one of the agents. Constraints \eqref{eq:nodeTimeWin} require that the time at which an agent visits a depot node $i \in \{s,s'\}$ or visits a target $\tau_i$ at a segment node $i \in V_{seg}$, corresponds to the time interval $[\underline{t}_i,\overline{t}_i]$ for that node. Constraints \eqref{eq:defSOCPlxy} captures the definition of the auxiliary variables $l_{e,k}^{xy}$.

Now, consider the big-$M$ constraints \eqref{eq:speedSOCPBigM} and \eqref{eq:coneSOCPBigM} for each edge $e=(i,j) \in E$, and each agent $k \in V_{agt}$. The big-$M$ values used here are $T$, and $R$, respectively. Constraints \eqref{eq:coneSOCPBigM} along with \eqref{eq:defSOCPlxy} and \eqref{eq:normSquaredSOCP} describe the condition that if $y_{e,k}=1$, then $l_{e,k} \geq \norm{p_{j,k}-p_{i,k}}{2}$ and if $y_{e,k}=0$, then $l_{e,k}$ is free to take any value. Note that to minimize the objective \eqref{eq:SOCPobj}, the condition: $l_{e,k} = \norm{p_{j,k}-p_{i,k}}{2}$ if $y_{e,k}=1$, and $l_{e,k}=0$ if $y_{e,k}=0$, must be satisfied. The time-feasibility constraints \eqref{eq:speedSOCPBigM} describe the condition that if $y_{e,k}=1$, then $l_{e,k} \leq v_{max}(t_{j,k}-t_{i,k})$ and if $y_{e,k}=0$, then no restrictions are placed on $t_{i,k}$ and $t_{j,k}$ (since $l_{e,k}=0$ in this case).

\vspace{1mm}
Although this MICP describes the MA-MT-TSP well, It is challenging to solve in practice. This is possibly due to the big-$M$ constraints \eqref{eq:speedSOCPBigM} and \eqref{eq:coneSOCPBigM}, which may lead to poor convex relaxations, as well as numerical instabilities, as noted by the authors in \cite{stieber2022}. In the next section, we present an alternative MICP based on the ideas in \cite{philip2024mixed} and \cite{marcucci2024shortest}, that finds the optimum for the MA-MT-TSP, much faster than the baseline. Prior to presenting this formulation, we restate the baseline as a nonconvex MINLP. This will aid us in proving that an optimal solution for our new MICP indeed provides an optimal solution for the MA-MT-TSP.

\section{A New MICP for the MA-MT-TSP}

\vspace{-2mm}

\subsection{Nonconvex MINLP Formulation for MA-MT-TSP}
This section details how the baseline MICP formulation for the MA-MT-TSP can be restated as a nonconvex MINLP. We refer to this program as the nonconvex formulation for simplicity. First, we define the decision variables. For each node $i \in V$ and agent $k \in V_{agt}$, we reuse the variable $t_{i,k}$ from the baseline. In addition, we also define $p_{i,k}$ as a new auxiliary variable. For each edge $e = (i,j) \in E$, we reuse the variables $y_{e,k}$, $l_{e,k}$, and $l_{e,k}^{xy}$ from the baseline. The $l_{e,k}$ variable is defined with the help of new real auxiliary variables $z_{e,k,p}$ and $z_{e,k,t}$ representing the products $y_{e,k}p_{i,k}$ and $y_{e,k}t_{i,k}$ respectively, and variables $z'_{e,k,p}$ and $z'_{e,k,t}$ representing $y_{e,k}p_{j,k}$ and $y_{e,k}t_{j,k}$ respectively.
Observe how in the baseline, $l_{e,k}$ was defined with the help of $\overline{l}_{e,k}$ and big-$M$ constraints instead. We now present the nonconvex formulation below.

\begin{align}
    \label{eq:ncxobj}
    &\min \;\; \sum_{k\in V_{agt}}\sum_{e\in E} l_{e,k} \\
    \intertext{subject to constraints \eqref{eq:depotFlowOut}, \eqref{eq:depotFlowIn}, \eqref{eq:targetFlowIn}, \eqref{eq:flowConservation}, \eqref{eq:nodeTimeWin},}  
    \label{eq:ncxnodepos}
    &p_{i,k} = \underline{p}_{i}+t_{i,k}v_{i}-\underline{t}_{i}v_{i}, \;\; \forall \; i \in V, \; k \in V_{agt}, \\
    \label{eq:ncxspeed}
    &l_{e,k} \leq v_{max}(z'_{e,k,t}-z_{e,k,t}), \; \; \forall \; e=(i,j) \in E, \; k \in V_{agt},
\end{align}
\begin{align}
    \label{eq:ncxlxy}
    &l_{e,k}^{xy}=(z'_{e,k,p}-z_{e,k,p}), \;\; \forall \; e=(i,j) \in E, \; k \in V_{agt},\\
    \label{eq:ncxnormSquared}
    &\norm{l_{e,k}^{xy}}{2}^2 \leq l_{e,k}^2, \;\; \forall \; e=(i,j) \in E, \; k \in V_{agt},\\
    \label{eq:ncxproducts}
    \begin{split}
        &z_{e,k,p}=y_{e,k}p_{i,k}, \; z_{e,k,t}=y_{e,k}t_{i,k}, \; z'_{e,k,p}=y_{e,k}p_{j,k},\\
        &z'_{e,k,t}=y_{e,k}t_{j,k}, \;\; \forall \; e=(i,j) \in E, \; k \in V_{agt}.
    \end{split}
\end{align}

\vspace{3mm}
This formulation shares the flow constraints \eqref{eq:depotFlowOut} to \eqref{eq:flowConservation} as well as the time of visit constraints \eqref{eq:nodeTimeWin} from the baseline. Constraints \eqref{eq:ncxnodepos} describe the variables $p_{i,k}$. Now, we explain the remaining constraints \eqref{eq:ncxspeed} to \eqref{eq:ncxproducts}. Note how apart from the binary requirement for the flow variables $y_{e,k}$, the nonconvexities in this program arise from the product of flow variables and node variables in \eqref{eq:ncxproducts}. These constraints help achieve the role satisfied by the big-$M$ constraints in the baseline. Note how when $y_{e,k}=1$, \eqref{eq:ncxproducts} gives $z_{e,k,p}=p_{i,k}$ and $z'_{e,k,p}=p_{j,k}$. Consequently, constraints \eqref{eq:ncxlxy} and \eqref{eq:ncxnormSquared} give $l_{e,k} \geq \norm{p_{j,k}-p_{i,k}}{2}$. However, when $y_{e,k}=0$, \eqref{eq:ncxproducts} gives $z_{e,k,p}=z'_{e,k,p}=(0,0)$ and consequently, $l_{e,k}=0$ from \eqref{eq:ncxlxy} and \eqref{eq:ncxnormSquared}. Like the baseline, in order to minimize the objective, the condition $l_{e,k}=\norm{p_{j,k}-p_{i,k}}{2}$ if $y_{e,k}=1$ and $l_{e,k}=0$ if $y_{e,k}=0$, must be satisfied. Now, consider the time-feasibility constraints \eqref{eq:ncxspeed}. When $y_{e,k}=1$, $z_{e,k,t}=t_{i,k}$ and $z'_{e,k,t}=t_{j,k}$ from \eqref{eq:ncxproducts}. Consequently, \eqref{eq:ncxspeed} gives $l_{e,k}\leq v_{max}(t_{j,k}-t_{i,k})$. However, when $y_{e,k}=0$, $t_{i,k}$ and $t_{j,k}$ are free to take any values since $l_{e,k}=z_{e,k,t}=z'_{e,k,t}=0$ in this case. Hence, the nonconvex formulation and the baseline MICP have the same optimal objective value. Moreover, an optimal solution for the MA-MT-TSP can be obtained from an optimal solution for the nonconvex formulation by retrieving the values taken by $y_{e,k}$, $t_{i,k}$, and $p_{i,k}$ variables.

\subsection{A New MICP Formulation for MA-MT-TSP}

In this section, we reformulate the nonconvex formulation, as a new MICP for the MA-MT-TSP. First, we define the decision variables. We discard the node variables $t_{i,k}$ and $p_{i,k}$ used in the nonconvex formulation. Additionally, we do not define a separate set of edge variables for each agent $k\in V_{agt}$. We then end up with a single set of decision variables $y_e$, $l_e$, $l_e^{xy}$, $z_{e,p}$, $z_{e,t}$, $z'_{e,p}$, $z'_{e,t}$ for each edge $e \in E$. Now we present the MICP.

\begin{align}
    \label{eq:newobj}
    &\min \;\; \sum_{e\in E}l_e \\
    \intertext{subject to constraints}
    \label{eq:newflowfromDepot}
    &\sum_{e \in E^{out}_s}y_e = \alpha, \\
    \label{eq:newflowtoDepot}
    &\sum_{e \in E^{in}_{s'}}y_e = \alpha, \\
     \label{eq:newnumPursuer}
    &1 \leq \alpha \leq m, \\
     \label{eq:newsegPerCluster}
    &\sum_{i \in C_u}\sum_{e \in E^{in}_i}y_e = 1, \;\; \forall \; u \in V_{tar}, \\
    \label{eq:newflowConservation}
    &\sum_{e\in E^{in}_i}(z'_{e,t},y_e) = \sum_{e\in E^{out}_i}(z_{e,t},y_e), \;\; \forall \; i \in V_{seg},
\end{align}
\begin{align}
    \label{eq:newtimei}
    & y_e\underline{t}_i \leq z_{e,t} \leq y_e\overline{t}_i, \;\; \forall \; e=(i,j) \in E,\\
    \label{eq:newtimej}
    &y_e\underline{t}_j \leq z'_{e,t} \leq y_e\overline{t}_j, \;\; \forall \; e=(i,j) \in E,\\
    \label{eq:newposi}
    &z_{e,p} = v_{i}z_{e,t} + y_e(\underline{p}_{i} - \underline{t}_iv_{i}), \;\; \forall \; e=(i,j) \in E,\\
    \label{eq:newposj}
    &z'_{e,p} = v_{j}z'_{e,t} + y_e(\underline{p}_{j}-\underline{t}_jv_{j}), \;\; \forall \; e=(i,j) \in E,\\
    \label{eq:newspeed}
    &l_e \leq v_{max}(z'_{e,t}-z_{e,t}), \;\; \forall \; e=(i,j) \in E, \\
    \label{eq:newlxy}
    &l_{e}^{xy} = (z'_{e,p}-z_{e,p}), \;\; \forall \; e=(i,j)\in E, \\
    \label{eq:newnorm}
    &\norm{l_{e}^{xy}}{2}^2 \leq l_e^2, \;\; \forall \; e=(i,j) \in E.
\end{align}

\vspace{2mm}
This formulation defines multiple agent tours without defining a separate set of edge variables for each agent. This is done by allowing an integer flow $\alpha$ where $1 \leq \alpha \leq m$, in and out of the depot, and restricting the flow in and out of the other nodes to be equal, and be at most 1. Recall that $m$ denotes the total number of agents. These conditions are described by the new flow constraints \eqref{eq:newflowfromDepot} to \eqref{eq:newflowConservation}, which effectively replaces \eqref{eq:depotFlowOut} to \eqref{eq:flowConservation} in the nonconvex formulation. Note that in \eqref{eq:newflowConservation}, an additional condition, $\sum_{e\in E_{i}^{in}}z'_{e,t}=\sum_{e\in E_{i}^{out}}z_{e,t}$ is added for each node $i\in V_{seg}$. This is to ensure that the incoming and outgoing edge for some segment node $i$ in a tour, coincide at the same time corresponding to that node. This additional condition is used since node variables for time are not introduced in this formulation.


\vspace{1mm}
Now, consider the remaining constraints \eqref{eq:newtimei} to \eqref{eq:newnorm}. Constraints \eqref{eq:newspeed} to \eqref{eq:newnorm} are similar to constraints \eqref{eq:ncxspeed} to \eqref{eq:ncxnormSquared} in the nonconvex formulation, with the only changes coming from not having a separate set of edge variables and corresponding constraints for each agent. Finally, constraints \eqref{eq:newtimei} to \eqref{eq:newposj} in the MICP replaces constraints \eqref{eq:nodeTimeWin}, \eqref{eq:ncxnodepos}, and \eqref{eq:ncxproducts} in the nonconvex formulation. The key idea here is as follows: For each $e=(i,j) \in E$, we define $(z_{e,p},z_{e,t}) \in y_eX_i$ and $(z'_{e,p},z'_{e,t}) \in y_eX_j$, where $(p_i,t_i)\in X_i$ represent the below two inequalities, similar to constraints \eqref{eq:nodeTimeWin} and \eqref{eq:ncxnodepos}, for some node $i \in V$.

\vspace{-1mm}
\begin{align}
    & \underline{t}_i \leq t_i \leq \overline{t}_i, \\
    & p_i = \underline{p}_i + t_iv_i - \underline{t}_iv_i.
\end{align}

\vspace{2mm}
Consequently, we let $(z_{e,p},z_{e,t}) \in X_i$ and $(z'_{e,p},z'_{e,t}) \in X_j$ when $y_e=1$, and become $(0,0,0)$ when $y_e=0$, without explicitly defining node variables $t_i$ and $p_i$, or multiplying them with the flow variables like in \eqref{eq:ncxproducts}. 

\vspace{1mm}
From an optimal solution to the MICP, optimal agent tours for the nonconvex formulation can be obtained by considering the subset of edges $E_{tour} := \{e \in E: y_e=1\}$ and finding all the $1 \leq \alpha \leq m$ vertex-disjoint paths starting at $s$ and ending at $s'$ within $E_{tour}$. This can be done using a graph search algorithm such as Depth-First Search (DFS). Since all the agents are homogeneous and share the same depot, any of the paths can be assigned to any agent. Let $E_{tour}^k \subseteq E_{tour}$ denote the path assigned to an agent $k \in V_{agt}$. Note that $E_{tour} = \bigcup_{k \in V_{agt}}E_{tour}^k$. Also note that it is possible $E_{tour}^k = \emptyset$ for some $k$ in the case where some of the agents are not assigned tours. Finally, note that for some $k \in V_{agt}$, $E_{tour}^k = \{e\in E: \; y_{e,k}=1\}$ for the nonconvex formulation. Hence, for each $e=(i,j) \in E_{tour}^k$, we set $y_{e,k}=1$, and set $(z_{e,k,p},z_{e,k,t}) =(p_{i,k},t_{i,k}) = (z_{e,p},z_{e,t})$ and $(z'_{e,k,p},z'_{e,k,t}) = (p_{j,k},t_{j,k}) = (z'_{e,p},z'_{e,t})$.


\subsection{Proof of Validity}
In this section, we will show the correctness of the MICP formulation by proving the following theorem.

\begin{theorem}
    The optimal value of the MICP formulation is equal to that of the nonconvex formulation for the MA-MT-TSP. Additionally, optimal agent tours for the nonconvex formulation can be recovered from an optimal solution to the MICP by following the procedure explained previously.
\end{theorem}

\begin{proof}
    Let $E_{tour}$ be the set of all edges selected at optimality by either of the two formulations. For the nonconvex formulation, $E_{tour}:=\{e\in E: \exists \; k \in V_{agt} \; \text{such that} \; y_{e,k}=1\}$, and for the MICP formulation, $E_{tour}:=\{e\in E: y_e=1\}$. The flow constraints in both formulations require that the edges in $E_{tour}$ form $1 \leq \alpha \leq m$ vertex-disjoint paths, that starts at $s$ and ends at $s'$, such that for each target $u \in V_{tar}$, exactly one node in $C_u$ is visited by one of the $\alpha$ paths. First, consider an edge $e \notin E_{tour}$. In the nonconvex formulation, for all $k \in V_{agt}$, $y_{e,k}=0$. Consequently, constraints \eqref{eq:ncxproducts} require $(z_{e,k,p},z_{e,k,t})=(z'_{e,k,p},z'_{e,k,t})=(0,0,0)$ for all $k\in V_{agt}$. Similarly, in the MICP formulation, $y_e=0$ leading to constraints \eqref{eq:newtimei} to \eqref{eq:newposj} requiring $(z_{e,p},z_{e,t})=(z'_{e,p},z'_{e,t})=(0,0,0)$. Hence, the cost addend is $\sum_{k\in V_{agt}}l_{e,k}=l_{e}=0$ for either formulation. Now, consider an edge $e=(i,j) \in E_{tour}$. In the nonconvex formulation, since edge $e$ is assigned to some unique agent $k^*\in V_{agt}$, $y_{e,k}=1$ when $k=k^*$ and $y_{e,k}=0$ for all other $k \in V_{agt} \setminus \{k^*\}$. Consequently, \eqref{eq:ncxproducts} requires $(z_{e,k^*,p},z_{e,k^*,t}) = (p_{i,k^*},t_{i,k^*})$ and $(z'_{e,k^*,p},z'_{e,k^*,t}) = (p_{j,k^*},t_{j,k^*})$. Additionally, \eqref{eq:nodeTimeWin} and \eqref{eq:ncxnodepos} require $(p_{i,k^*},t_{i,k^*}) \in X_i$ and $(p_{j,k^*},t_{j,k^*}) \in X_j$. These two requirements, with the help of flow conservation, \eqref{eq:flowConservation} reads $(z_{e,k^*,p},z_{e,k^*,t})\in X_i$, $(z'_{e,k^*,p},z'_{e,k^*,t}) \in X_j$, and for adjacent edges $e=(i,j)$ and $f=(j,k)$ in $E_{tour}$, $(z'_{e,k^*,p},z'_{e,k^*,t})=(z_{f,k^*,p},z_{f,k^*,t})$. Similarly, in the MICP formulation, $y_e=1$ and therefore, constraints \eqref{eq:newtimei} to \eqref{eq:newposj} require $(z_{e,p},z_{e,t}) \in X_i$, $(z'_{e,p},z'_{e,t}) \in X_j$. Additionally, the new requirement in \eqref{eq:newflowConservation} ensures that for adjacent edges $e=(i,j)$ and $f=(j,k)$ in $E_{tour}$, $(z'_{e,p},z'_{e,t})=(z_{f,p},z_{f,t})$. Finally, the time-feasibility constraints in both the formulations \eqref{eq:ncxspeed} and \eqref{eq:newspeed} have the same form. Therefore, corresponding to an edge $e \in E_{tour}$, the cost addend in the nonconvex formulation is $\sum_{k\in V_{agt}}l_{e,k}=l_{e,k^*}=\norm{z'_{e,k^*,p}-z_{e,k^*,p}}{2}$ which is the same as the cost addend $l_e=\norm{z'_{e,p}-z_{e,p}}{2}$ by the MICP formulation. Since the $E_{tour}$ corresponding to an optimal solution for one formulation, when used in the other formulation, results in a feasible solution with the same cost, both formulations share the same optimal value. Moreover, from an optimal solution to the MICP, optimal agent tours for the nonconvex formulation can be obtained by following the previously described procedure.
\end{proof}

\section{Numerical Results}
\subsection{Test Settings and Instance Generation}
All the tests were run on a laptop with an Intel Core I7-7700HQ 2.80GHz CPU, and 16GB RAM. The implementation was in Python 3.11.6, and both the baseline MICP formulation (MICP-Baseline) and the new MICP formulation (MICP) were solved using Gurobi 10.0.3 optimizer \cite{gurobi}. All the Gurobi parameters were set to their default values, except for \emph{TimeLimit} \footnote{Limits the total time expended (in seconds).}, which was set to 1800.

\vspace{1mm}
A total of 80 instances were generated, with 20 instances each for 5, 10, 15, and 20 targets. Each instance was defined by the number of targets $n$, a square area of fixed size $S=100 \; units$ (with diagonal length $R = \sqrt{2}S$), a fixed time horizon $T = 150 \; secs$, the depot location fixed at the center $(0,0)$ of the square, the maximum speed of the agents $v_{max} = 4 \;\; units/sec$, and finally, a set of randomly generated piecewise-linear trajectories corresponding to the $n$ targets such that each target has a constant speed within $[0.5,1] \; unit/sec$, and is confined within the square area.

\vspace{1mm}
For each instance, we conducted experiments where we varied two test parameters: (a) the total time window duration for each target, which is the sum of durations of all the time windows associated with the target, and (b) the number of agents $m$. The total time window duration was varied to be 20, 40, and 60 $secs$, and the number of agents were varied to be from 1 to 5. Each target was assigned two time windows of equal duration in all the experiments. The time windows were selected such that for all generated instances, a feasible solution exists even when restricting the number of agents to 1. This was done by randomly choosing a sequence in which all the targets are visited by a single agent, and then finding the quickest tour for that sequence by fixing the speed of the agent to be $v_{max}$. If the time to complete the tour exceeded $T$, the process was repeated with another sequence. Otherwise, for all choices of total time window durations, two non-intersecting time windows of half the total duration were defined for each target such that one of the time windows contains the time at which the target was visited in the tour. 

\begin{figure}[ht!]
    \centering
    \includegraphics[width=\linewidth]{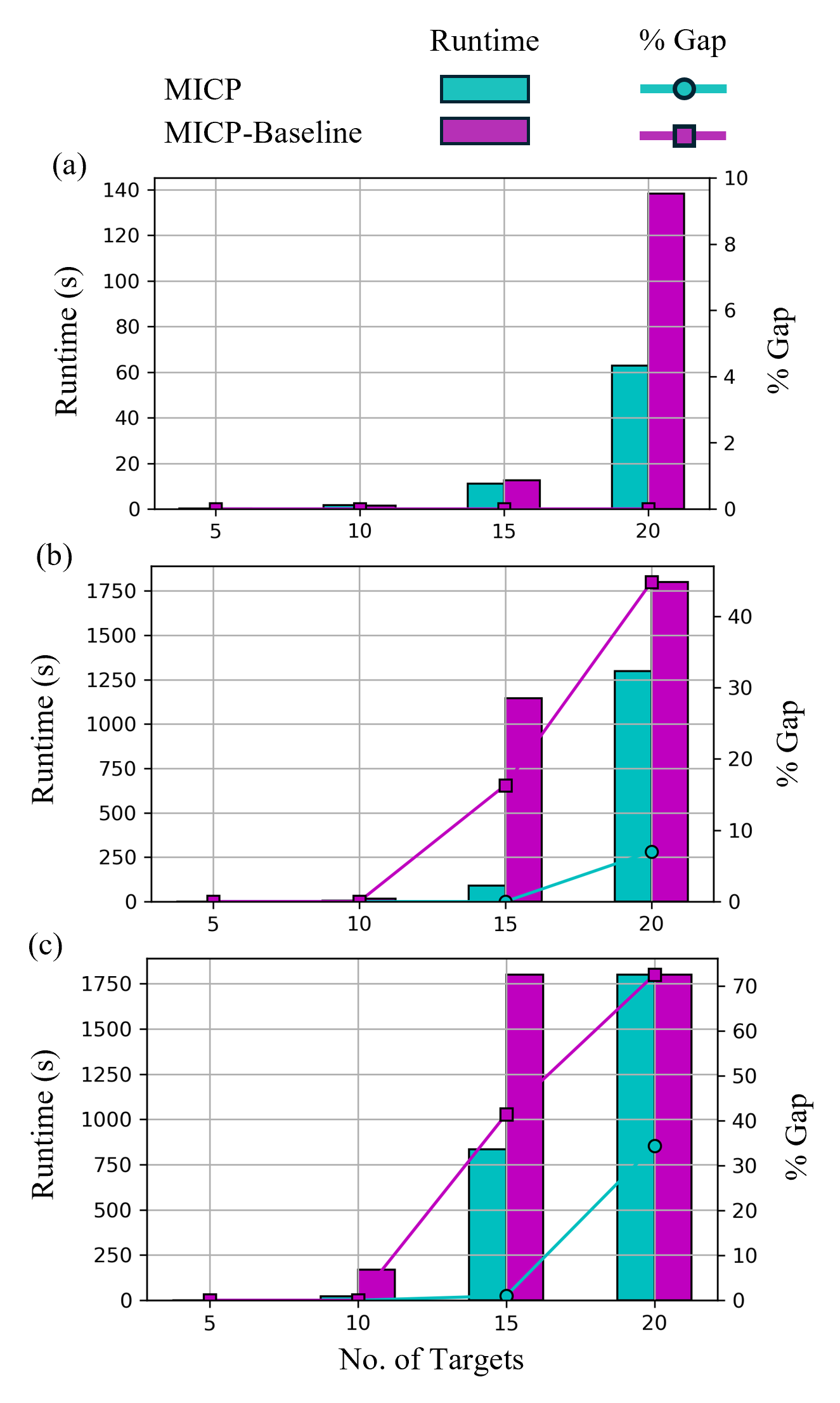}
    \caption{Numerical results comparing the \% Gap and runtime for MICP-Baseline and MICP, when the number of agents is fixed at 1, and the total time window duration is varied to be 20 (a), 40 (b), and 60 (c) $secs$. The MICP scales significantly better than MICP-Baseline with larger time window durations and more targets. For 10 targets in (c) and 15 targets in (b), the MICP shows a runtime improvement of two orders of magnitude. It also shows a \% Gap improvement of more than 15 for 15 targets in (b). Similarly, for 15 targets in (c) and 20 targets in (b) and (c), the MICP runs up to 1000 seconds faster, while also providing a \% Gap improvement within a 35-40 range.}
    \label{fig:expVaryTw}
    \vspace{-2mm}
\end{figure}

\vspace{1mm}
To evaluate the MICP-Baseline and MICP formulations, we use \% Gap, and runtime, which we will now explain. For a given total time window duration, number of agents, and formulation of choice, the solver is first run on all the 20 instances corresponding to a given number of targets. The optimality gap value from the solver for an instance is defined as $\frac{|c_f-c_{lb}|}{|c_f|}\times 100$, where $c_f$ is the feasible (primal) objective, and $c_{lb}$ is the lower bound (dual) objective. \% Gap denotes the average of the smallest gap values output by the solver within the time limit for all these instances, and runtime denotes the average of the run-times output by the solver for all these instances.

\subsection{Varying the Total Time Window Duration}
In this section, we present the results of experiments where the total time window duration was varied. The number of agents were fixed at 1, and all the instances were solved using both formulations for the varying durations (20, 40, and 60 $secs$). The results of this experiment are illustrated in Fig.~\ref{fig:expVaryTw}, with panels (a), (b), and (c) corresponding to durations 20, 40, and 60, respectively. We observe that for both the formulations, the problem becomes more challenging to solve as the number of targets increases. Moreover, this difficulty becomes more prominent as the total time window duration increases. However, we observe that the MICP scales significantly better than MICP-Baseline against larger number of targets and bigger time windows. This can be seen especially in the case of 15 targets where the \% Gap always converge to 0 for the MICP, and its runtime increases noticeably only with the largest time window duration of 60, as opposed to the MICP-Baseline whose \% Gap and runtime grows dramatically with increasing time window durations. Note how at 20 targets, the problem is challenging for both formulations. However, MICP still significantly outperforms MICP-Baseline in terms of runtime or \% Gap for all the time window durations considered.



\subsection{Varying the Number of Agents}

\begin{figure}[ht!]
    \centering
    \includegraphics[width=\linewidth]{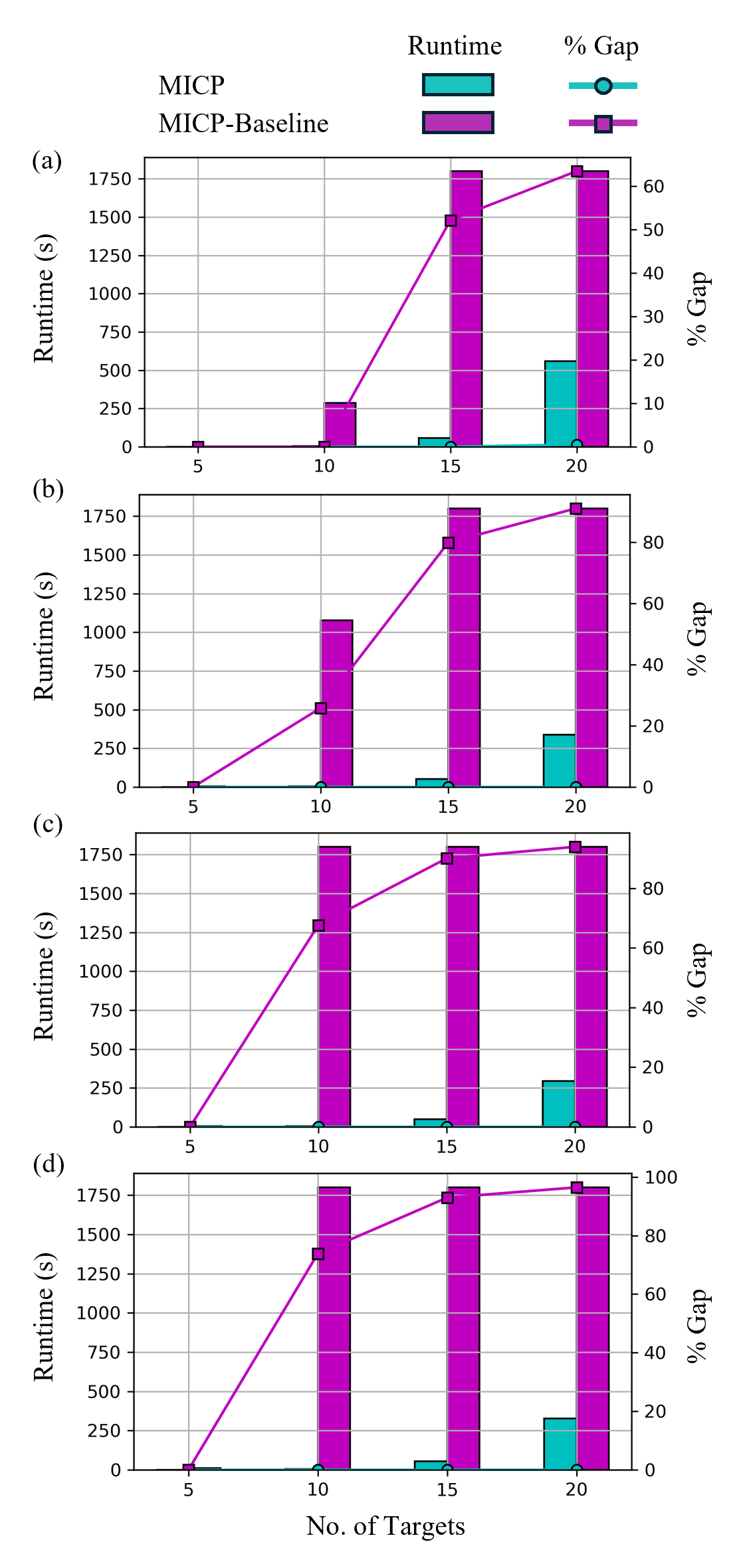}
    \caption{Numerical results comparing the \% Gap and runtime for MICP-Baseline and MICP, when the total time window duration is fixed at 40, and the number of agents is varied to be 2 (a), 3 (b), 4 (c), and 5 (d). The MICP significantly outperforms MICP-Baseline for all number of targets and agents. Note how for 15 targets, MICP runs two orders of magnitude faster, while giving a \% Gap improvement of more than 50 for 2 agents to more than 90 for 5 agents. Similarly, for 20 targets, the MICP runs one-order of magnitude faster, while giving a \% Gap improvement of 60 for 2 agents to more than 90 for 5 agents.}
    \label{fig:expVaryPursuers}
    \vspace{-5mm}
\end{figure}

In this section, we present the results of experiments where the number of agents was varied. The total time window duration was fixed at 40 $secs$, and all the instances were solved using both formulations for 2, 3, 4, and 5 agents. The results of this experiment are illustrated in Fig.~\ref{fig:expVaryPursuers}, with panels (a), (b), (c), and (d) corresponding to 2, 3, 4, and 5 agents, respectively. Recall that the 1 agent case was already presented in Fig.~\ref{fig:expVaryTw} (b). Hence, it is not repeated in Fig.~\ref{fig:expVaryPursuers}. Like before, the \% Gap and runtime worsen with more targets for both formulations. With additional agents, this effect is further pronounced for the MICP-Baseline. However, for the MICP, we observe an improvement in results in this case. From Fig.~\ref{fig:expVaryPursuers}, we observe that the MICP-Baseline struggles significantly with 2 or more agents for 15-target instances, and with 4 or more agents for 10-target instances. On average, we find the solver reaches the time limit in these cases, with a remaining \% Gap of more than 60. However, for the MICP, the \% Gap always converges to 0. Moreover, we observe a significant improvement in runtime, with up to two orders of magnitude for 10 and 15 targets, and one order of magnitude for 20 targets. This can be attributed to the MICP not using a separate set of decision variables for each agent and to the improved convexification process it uses, which does not rely on big-$M$ constraints.

\section{Conclusion And Future Work}
In this paper, we presented a new Mixed-Integer Conic Program formulation that finds the optimum for the Multi-Agent Moving-Target Traveling Salesman Problem. We considered the case where the agents are homogeneous and share the same depot, and each target moves along piecewise-linear segments and have more than one associated time window. We proved the validity of our formulation, and provided numerical results to corroborate its performance. We showed how our MICP outperforms the state-of-the-art MICP, both in terms of runtime and optimality gap across various experiments. For future work, we plan on extending our formulation to handle a more general case of the problem with multiple depots and heterogeneous agents.





\section*{Acknowledgment}
This material is based upon work supported by the National Science Foundation under Grant 2120219 and Grant 2120529. Any opinions, findings, and conclusions or recommendations expressed in this material are those of the author(s) and do not necessarily reflect the views of the National Science Foundation.



\bibliographystyle{plain}
\bibliography{references}

\end{document}